\documentclass[sigconf]{acmart}

\copyrightyear{2023}
\acmYear{2023}
\setcopyright{acmlicensed}
\acmConference[MM '23] {Proceedings of the 31st ACM International Conference on Multimedia}{October 29--November 3, 2023}{Ottawa, ON, Canada.}
\acmBooktitle{Proceedings of the 31st ACM International Conference on Multimedia (MM '23), October 29--November 3, 2023, Ottawa, ON, Canada}
\acmPrice{15.00}
\acmISBN{979-8-4007-0108-5/23/10}
\acmDOI{10.1145/3581783.3612356}

\usepackage{graphicx}
\usepackage{amsmath}
\usepackage{balance}
\usepackage{booktabs}
\usepackage{multirow}
\usepackage{color}
\usepackage{subfigure}
\usepackage{makecell}
\graphicspath{{figures/}}
\graphicspath{{Image/}}

\begin{document}


\title[P2I-NET: Mapping Camera Pose to Image via Adversarial Learning  for New View Synthesis \\ in Real Indoor Environments]{P2I-NET: Mapping Camera Pose to Image via Adversarial Learning  for New View Synthesis in Real Indoor Environmentse}


\author{Xujie Kang }
\affiliation{%
  \institution{College of Electronics and
Information Engineering, Shenzhen University, China}
  \streetaddress{No. 3688, Nanhai Avenue, Nanshan District, Shenzhen City, Guangdong Province}
  \postcode{518000}
  \country{}
}
\email{kangxj@szu.edu.cn}

\author{Kangling Liu }
\affiliation{%
  \institution{Peng Cheng Laboratory, Shenzhen, China }
  \country{}
  }
\email{max.liu.426@gmail.com}

\author{Jiang Duan }
\authornote{Also with fotor.com}
\affiliation{%
  \institution{ School of Computing and Artificial Intelligence, Southwestern University of Finance and Economics, China}
  \country{}
  }
\email{duanj_t@swufe.edu.cn}

\author{ Yuanhao Gong}
\affiliation{%
  \institution{College of Electronics and
Information Engineering, Shenzhen University, China}
\country{}
  }
\email{gong@szu.edu.cn}

\author{Guoping Qiu }
\authornote{The corresponding author}
\authornote{Also with fotor.com and the School of Computer Science, The University of Nottingham, UK (guoping.qiu@nottingham.ac.uk)}
\affiliation{%
  \institution{College of Electronics and
Information Engineering, Shenzhen
University, China}
\country{}
   }
\email{qiu@szu.edu.cn}


\begin{abstract}
 Given a new $6DoF$ camera pose in an indoor environment, we study the challenging problem of predicting the view from that pose based on a set of reference RGBD views. Existing explicit or implicit 3D geometry construction methods are computationally expensive while those based on learning have predominantly focused on isolated views of object categories with regular geometric structure. Differing from the traditional \textit{render-inpaint} approach to new view synthesis in the real indoor environment, we propose a conditional generative adversarial neural network (P2I-NET) to  directly predict the new view from the given pose. P2I-NET learns the conditional distribution of the images of the environment for establishing the correspondence between the camera pose and its view of the environment, and achieves this through a number of innovative designs in its architecture and training lost function. Two auxiliary discriminator constraints are introduced for enforcing the consistency between the pose of the generated image and that of the corresponding real world image in both the latent feature space and the real world pose space. Additionally a deep convolutional neural network (CNN) is introduced to further reinforce this consistency in the pixel space. We have performed extensive new view synthesis experiments on real indoor datasets. Results show that P2I-NET has superior performance against a number of NeRF based strong baseline models. In particular, we show that P2I-NET is 40 to 100 times faster than these competitor techniques while synthesising similar quality images. Furthermore, we contribute a new publicly available indoor environment dataset containing 22 high resolution RGBD videos where each frame also has accurate camera pose parameters. 
\end{abstract}






\maketitle

\section{Introduction}
\label{sec:intro}
With the emergence of technologies such as Generative Adversarial Networks (GAN) \cite{goodfellow2020generative}, diffusion models \cite{rombach2022high}, and text-to-image and image-guided image generation \cite{zhang2023adding}, image synthesis has seen remarkable progress in recent years. However, accurately controlling the generation of real scene images based on 6DoF camera pose remains a significant challenge because the camera pose contains very little information. 
Computer graphics traditionally uses standard rendering techniques to obtain realistic images from a given camera perspective. However, this approach requires explicit simulation of various aspects of the scene, such as geometry, materials, and light transmission, making building and editing virtual scene maps expensive and time-consuming. By transforming graphics rendering into a data-driven mode, the image rendering process based on camera pose can be greatly simplified. The popular NeRF \cite{r1} technology uses a multilayer perceptron (MLP) to implicitly construct the 3D radiance field of the scene and synthesize an image of a given camera perspective through a fully differentiable radiance field and volume rendering method, achieving impressive results. However, this learning reasoning model is often complicated and time-consuming. Follow-up NeRF research works \cite{barron2021mip} \cite{muller2022instant} mainly focus on image generation quality and rendering speed. However, most of these methods were tested on synthetic data and imposed significant demand on the collection of experimental datasets.


Suppose we are using a camera to capture an image set $\{I\}$ of a static scene.  $\{I\}$  represents discrete samples of the image distribution $p(I)$ of the scene. Each image in the scene is determined by the camera pose $y$ and other imaging parameters such as object materials, geometric structures, lighting conditions of the scene, and camera sensor properties. For a given scene and a camera which are fixed, then the only factor that determines the image is the camera pose $y$. The image distribution can be written as $p(I|y)$. Therefore, if we can estimate the continuous image distribution $p(I|y)$ of a specific scene by the discrete set of image samples $\{I\}$ of the scene, then we can sample the distribution $p(I|y)$ by providing discrete pose values $y_i$ to synthesis images viewed from $y_i$, i.e, the new view from the new pose $y_i$ is $I_i = p(I|y_i)$. 
Therefore, it is clear that new view synthesise for a given static scene can be achieved through estimating the distribution $p(I|y)$. It is known that the generator of a GAN is capable of capturing data distribution \cite{goodfellow2020generative}. In this paper, we have developed a CGAN network called the camera pose to image mapping neural network (P2I-NET) to achieve new view synthesis in real indoor environments. Main contributions of the paper are: 

(1) A camera pose to image mapping neural network (P2I-NET) has been successfully developed for new view synthesis in real indoor environments. The P2I-NET learns the conditional distribution of the images of the environment thus establishing the correspondence between the camera pose and its view of the environment. The innovative features of the P2I-NET architecture and its training loss function include two discriminator auxiliary constraints, one in a high dimensional latent feature space and the other in the low-dimensional real world pose space, to force the consistency between the poses of the generated image and that of the real world image. In addition, an enhancement subnet is introduced to reinforce this consistency in the pixel space.

(2) A new dataset suitable for researching new view synthesise in the real environments and related topics such as 3D environment construction has been developed and will be made publicly available. The new Camera Pose to Virtual Video (CP2V$^2$) dataset contains 22 high resolution RGBD videos (a total of 55,000 frames) which were taken from two indoor environments by attaching an RGBD camera to a robotic arm. Amongst other useful information, each frame also contains their accurate camera pose parameters.  

(3) We have performed extensive new view synthesis experiments using the new CP2V$^2$ dataset and the publicly available 7 Scenes dataset. We compare P2I-NET with a number of NeRF based top performing competitor techniques and show that P2I-NET has a comparable performance. In particular, we show that P2I-NET is 40 to 100 times faster than these SOTA techniques for new view synthesis while synthesising similar quality images.

\section{Related Work}
\label{RelativeWork}
 Camera pose estimation from image is a crucial task in computer vision. Recent years have seen the emergence of learning-based methods that employ CNNs such as VGG\cite{r3} or ResNet\cite{r4} to model the hidden correspondence between images and camera poses. This paper study the inverse problem - estimating image from camera pose.  
 
\textbf{Pose to image generation: explicit models}. In terms of image synthesis based on camera pose, several works exploit generative 3D models for 3D-aware image synthesis\cite{r91,r92,r93}, which aim at generating 3D representations and explicitly models the image formation process. 
The work \cite{2018DeepVoxels} achieved RGB and depth image synthesis from 2D image datasets through learning occlusion aware projections from 3D latent feature to 2D in an unsupervised manner. The works \cite{liu2020disentangled} and \cite{2018FaceID} learned to generate images by controlling camera poses using camera pose annotations or images captured from multiple viewpoints. While the aforementioned methods show impressive results, they are restricted to isolated views of object categories from a synthetic dataset. The authors of \cite{novotny2019perspectivenet} proposed an alternative that uses both camera intrinsics and extrinsics to transfer pixels from referenced RGBD views for new view synthesis, rather than directly establishing the relationship between camera extrinsics and the RGBD images. Although they also consider new view synthesis for real indoor environments, their so-called \textit{render-inpaint} approach is very different from our direct learning method. 


\textbf{Pose to image generation: implicit neural representations}. Recently, a promising direction is encoding scenes in the weights of an multilayer perception $(MLP)$ that directly maps from a 3D spatial location to an implicit representation of the shape or other graphics functions, such as the signed distance \cite{r85}, textured materials \cite{r86,r87,r88,r89}, occupancy fields\cite{r102,r103} and illumination values\cite{r90}. Many methods construct models using 3D geometry as supervision information\cite{r97,r98} or assume 3D information as input \cite{r99,r100}. 
The Generative Query Network (GQN) \cite{2018Neural} can render new viewpoints given a latent encoding of the scene and a novel viewpoint, however it has only been tested on synthetic environments but not real world setups. The popular NeRF\cite{r1} technology uses a MLP to implicitly construct the 3D radiation field of the scene, and synthesizes an image of a given camera perspective through a fully differentiable radiation field and volume rendering method, and has achieved impressive results. Mip-NeRF \cite{barron2021mip}, representing scenes at a continuous-valued scale and by effectively rendering conical frustums rather than rays, reduces objectionable aliasing artifact and significantly improves NeRF\cite{r1}'s ability to represent fine details. Although achieving impressive results, almost all of these works focus on single objects or small-scale scenes and require multi-view training data acquired from certain directions and poses. 
In contrast, our work builds a CGAN to directly map camera poses to their images of the real complex environments. 

 \textbf{Pose to image generation: direct mapping}. The only work we can find that is somewhat similar to ours is RGBD-GAN \cite{2019RGBD} which uses the camera parameters as conditions to control RGBD image generation which uses an explicit 3D consistency loss to ensure two generated RGBD images with different camera parameters to be consistent with the 3D world. However, it can only generate a different viewpoint for a given input view and cannot establish the correspondence between arbitrary poses and their viewpoints of a real environment. 

\section{P2I-NET}

\subsection{Rationale}
Suppose we are using a camera to capture a set of images in a scene, $I\in \{ I_1,I_2,I_3…I_n\}$, where $I$ represents discrete samples of the image distribution $p(I)$ of the scene. Each image in the scene is determined by the camera pose $y$, camera parameters $C$, the objects and their geometries $R$, and the lighting conditions $L$. Therefore, the image distribution in the scene can be expressed as $p(I|y,R,L,C)$. Once the scene is determined, i.e., object material, geometric structure, lighting conditions, and camera are fixed, the only factor that determines the image is the camera pose $y$. The image distribution can be approximated as $ p(I|y,R,L,C)=\int \int \int p(I|y,R,L,C)dR dL dC = p(I|y)$. Therefore, if we can estimate the continuous image distribution $p(I|y)$ of a specific scene based on the priors given by the discrete set of image samples $I$ of the scene, then we can sample the distribution $p(I|y)$ by providing discrete pose values. Given a new pose $y'_i$, then the image of the scene viewed from it can be obtained $I'_i = p(I|y'_i)$. Based on the above reasoning, new view synthesise for a given static scene can be achieved through estimating the distribution $p(I|y)$. It is known that the generator of a generative adversarial network (GAN) is capable of capturing the data distribution \cite{goodfellow2020generative}. It is based on this rationale, we design a CGAN network called the camera pose to image mapping neural network (P2I-NET) to achieve new view synthesis in real indoor environments.   

\begin{figure*}[h] 
\centering 
\includegraphics[width=\textwidth,height=9.0cm]{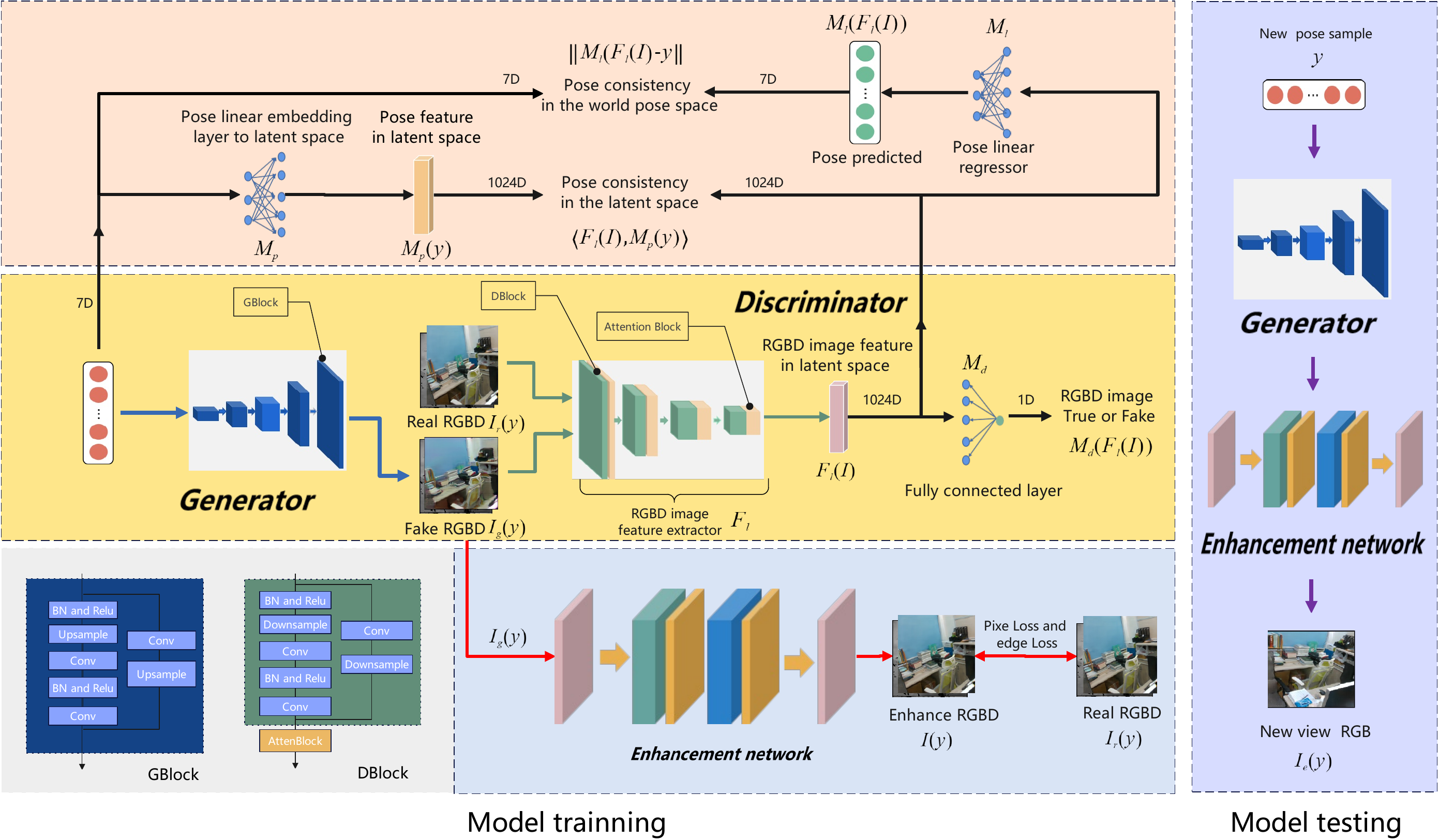}
\caption{The architecture of the camera pose to image mapping neural network (P2I-NET).In trainning, the network employs adversarial training to enforce consistency between pose and image in both the world pose space and the image latent feature space, establishing an intrinsic correspondence between pose and image. The generated image $I_g(y)$ is further constrained in the image pixel space to reinforce consistency between pose and image by the enhancement network. Ultimately, the P2I-NET network can generalize to generate high-quality RGB images $I_e(y)$ from new viewpoints in testing.} 
\label{RGBDPoseGAN} 
\end{figure*}

\subsection {Architecture Design}

\textbf{Overview}: Fig. \ref{RGBDPoseGAN} shows the architecture of the camera pose to image mapping neural network (P2I-NET). For a trained P2I-NET which would have established the intrinsic correspondence relationship between the poses and the images, new view synthesis is very simple and is achieved by inputting the 6DoF camera pose $y$ of a view nearby the train data to the generator network $G$, which will produce an image viewed from the pose $I_g(y)=G(y)$. To improve the visual quality of the reconstructed image, $I_g(y)$ is passed through an image enhancement deep CNN ($ENET$), and the final new view image is $I_e(y)=ENET(I_g(y))$. Most of our contributions are in the design of an effective training procedure which we will now describe in detail. 


\textbf{The Generator}: Unlike the conventional CGAN where the network input consists of a noise signal and a condition signal, the input to the Generator ($G$) in P2I-NET has no noise signal but rather it only has the condition signal which is a 7-dimensional camera pose information, that is $G$ generates the image viewed from the input pose directly. Suppose we are generating a $256\times256$ $RGB$ image, then $G$ maps a 7-dimensional vector to a $256\times256\times3 = 196608$ dimensional vector. The dimensionality of the output vector is $28086$ times higher that of the input, such a mapping is obviously extremely under constrained. At first glance, such mapping may seem impossible. As we shall show later, with deep neural network and adversarial learning, it is possible to to successfully achieve such a challenging mapping. This work further demonstrates the power of deep neural network and adversarial learning. 

\textbf{The Discriminator}: The discriminator $D$ performs adversarial learning \cite{goodfellow2020generative} to force the distribution of the generator output $P(G(y))$, to be as close as possible to the distribution of the real views of the environment $P(I_r(y))$, where $I_r(y)$ is the real environment image viewed from the pose $y$. The input to $D$ is $I_r(y)$ and $I_g(y)$, and the learning task of $D$ is to distinguish the two types of input samples as belonging to the $True$ class .
The architecture of the discriminator network is a CNN with attention mechanisms where each convolutional block is followed by an attention block. These attention blocks enable the discriminator to focus on the key features present in the input images. The number of feature channels in the hidden layer is made to be proportional to the number of channels and the resolution of the input image.

\textbf{Discriminator Auxiliary Constraints}: As discussed previously, P2I-NET directly maps a $7d$ pose parameters to an image, the size of which can be $256\times256\times3$ or higher. The mapping is clearly severely under constrained. Only enforcing the distribution of the generated outputs and that of the real environment views to be the same is not sufficient. Extra constraints are needed to ensure P2I-NET successfully learn such a difficult mapping, and we introduce two discriminator auxiliary constraints. 
Specifically, we use the output of the last convolutional layer which is $1024d$ vector (other dimensionality is possible but in this paper we use $1024d$) to construct two constraining conditions. Let $F_l(I)$ be the output of the last convolutional layer of $D$, $I\in\{I_g, I_r\}$.  $F_l(I)$ can be regarded as a latent representation of the input image and contains information about the image and its view points within the environment. We first enforce pose consistency in the latent space by first mapping the input $7d$ pose parameters $y$ to a $1024d$ vector using a simple mapping network $M_p$, $F_p(y)= M_p(y)$ and then forcing $F_l(I) = F_p(y)$, $I\in\{I_g, I_r\}$. We also enforce pose consistency in the world space by first mapping the latent feature $F_l(I)$ to a $7d$ camera pose parameters by another mapping network $M_l$, $y_l(I) = M_l(F_l(I)$ and forcing $y_l(I) = y$, $I\in\{I_g, I_r\}$. We will show these auxiliary constraints are both important in ensuring the successful training of the P2I-NET.

\textbf{Enhancement network (ENET)}: The generator $D$ network has a simple structure and its main task is to ensure its outputs to have data distribution and content structure consistencies with the real images. However, it does not specifically focus on the reconstruction of image details which may result the reconstructed image having poor visual quality. 
To address this issue, we further apply pose consistency constraint in the pixel space between the generated image $I_g(y)$ and the ground truth image $(I_r(y))$. 
However, directly imposing image pixel space consistency constraint within the GAN structure may potentially disrupt the adversarial training process. 
We therefore separate the image generation and quality enhancement processes into different networks which are trained separately (see training procedures in Section \ref{TrainingSec}). In this way, the burden on the generator network is reduced so that it can focus on learning data distribution and image structure consistencies. 

\textbf{Use of depth data}: RGBD cameras which capture the RGB as well as depth are readily available. By incorporating depth data, which contains geometric structure information, we can enhance the performance of P2I-NET. Therefore, the P2I-NET's input images have 4 channels.  

\subsection{Training}
\label{TrainingSec}
Training of P2I-NET is divided into two phases. The first and the most important, is training the generator and the discriminator based on adversarial learning. After the first phase training is completed, the generator is fixed, and the second phase trains the enhancement network ENET only. As discussed previously and with reference to Fig.\ref{RGBDPoseGAN}, in addition to these 3 major modules, there are two auxiliary constraint modules, $M_l$ for mapping the latent space features to the camera pose parameters in the real world space, and $M_p$ for mapping the real world camera parameters to the latent feature space. One of the challenges of training the $M_l$ is when the input to $M_l$ is the latent feature of the generated image $I_g$, for which there is no corresponding pose ground truth. We now explain how to construct the loss function for training the P2I-NET. 

\textbf{Discriminator Training}: The loss function for training the discriminator $D$ comes from three parts: image authenticity discrimination, pose consistency in the latent feature space and pose consistency in the real world coordinate space. 
We first construct a conditional projection discriminator \cite{r71} by combining the authenticity probability of the image with the cosine similarity measurement between the latent feature vectors: 

 \begin{equation}
 \label{eq_D_proj}
D_{pro}(I,y)=M_d(F_l(I)) +k_1〈F_l(I),M_p(y)〉
 \end{equation}
 
where $M_d$ is a mapping network which maps the latent feature $F_l(I))$ to a decision probability value for determining if the input sample is a real image or a generated image, $k_1$ is a weighting constant.  
Applying KL divergence and hinge loss to describe the difference between the real data distribution and the generated data distribution, the objective function of authenticity discrimination under pose consistency constraint in the the high-dimensional latent feature space can be written as:
\begin{equation}
\label{L_proj}
\begin{aligned}
L_{pro}=E_{I_r\sim P_{I_r} }\lbrack min(0,-1+D_{pro}(I_r,y)\rbrack+\\
E_{y\sim P_y }\lbrack min(0,-1-D_{pro}(I_g(y),y)\rbrack
\end{aligned}
\end{equation}

To enforce pose consistency in the real world space is more challenging. In the early stage of the training process, the difference between the generated image and the real image will be large, therefore, the pose parameters estimated from the generated images will have a larger discrepancy from that estimated from the real images. Therefore, we should allow a certain error range $\gamma$ between the pose estimated from the generated images and that estimated from the real images. This error range $\gamma$ should gradually decrease as the network training progresses. 
When the input is a real sample, its pose is known and we force the pose parameters estimated from the input to be exactly the same as the actual pose. When the input is a generated image, the estimated pose should be allowed to have a certain range of difference from that estimated from the real image. Therefore, for the pose consistency constraint in the real world space, the cost function for the real sample image and the generated sample image can be written respectively as:


\begin{equation}
L_{PE-I_r}=E_{I_r\sim P_{I_r} }  || M_l (F_l (I_r))  - y||\\
\end{equation}
\begin{equation}
\begin{aligned}
L_{PE-I_g(y)}=E_{y\sim P_y } max(||M_l (F_l (I_g(y)))- M_l (F_l (I_r))||-\gamma,0)
\end{aligned}
\end{equation}

where $M_l(F_l (I_g(y)))$ is the pose estimated from the generated image $I_g(y)$, $M_l (F_l (I_r))$ is the pose estimated from the real image $I_r$, and $\gamma$ is the error range, which is related to the content discrepancies expressed by $DIFF=|| M_d(F_l(I_r))  - M_d(F_l(I_g(y)))||$ . We set $\gamma = DIFF\times M_l(F_l (I_r))$, this will make the error range $\gamma$ gradually decreases as the network training progresses because $DIFF$ will become smaller as training progresses and the generated images getting more similar to the real ones. Finally, the total cost function of the discriminator $D$ is:
\begin{equation}
L_D=L_{pro}+k_2(L_{PE-I_r}+L_{PE-I_g(y)})
\label{equation D}
\end{equation}
where $k_2$ is a weighting constant.

\textbf{Generator Training}: 
The cost function for the Generator $G$ training can be written as:
\begin{equation}
\begin{aligned}
L_G=- E_{y\sim p_y } [D_{pro}(I_g(y),y)]+E_{y\sim p_y} ||M_l (F_l (I_g(y)))  - y||
\label{equation G}
\end{aligned}
\end{equation}

where $-E_{y\sim p_y } [D_{pro}(I_g(y),y)]$ forces the distribution of the generated images to be consistent with the distribution of the real images, $E_{y\sim p_y} ||M_l (F_l (I_g(y)))  - y||$ enforces the consistency between the pose estimated from the generated image and that of the real image. Starting from a camera pose as input, the generator produces an image and we force the pose of the generated image to be the same as the input pose, in this way the network form as closed loop. 

\textbf{ENET Training}: 
After training of the Generator and the Discriminator through optimizing the lost functions in (\ref{equation G}) and (\ref{equation D}) is completed, the Generator is fixed. We then start the second training phase by training the image enhancement network (ENET). 
ENET is a standard convolutional neural network and we construct pixel lost and an edge lost using the $L_2$ norm:
\begin{equation}
\label{ENET_LOST}
\begin{aligned}
L_{ENET}= E_{y\sim p_y } ||I_e(y) - I_r(y))|| + \\    k_3 E_{y\sim p_y}||F_{edge}(I_e(y))-F_{edge}(I_r(y))||
\end{aligned}
\end{equation}
where $I_e(y)$ is the enhancement network's output which is also the final output of the P2I-NET, $I_r(y)$ is the real environment image (ground truth) viewed from the camera pose $y$, $F_{edge}(I_e(y))$ and $F_{edge}(I_r(y))$ are the edge map images of the ENET output and the real environment image (ground truth). The edge map can be computed using any edge operators and we use the simple Roberts edge operator \cite{2002Woods}. $k_3$ is a weighting constant to balance the two terms. With this lost function, we ensure that the reconstruction is consistent with the real sample image in terms of pixel intensity (color) as well as in object contours ( structure). 

\section{Experimental Results}


\subsection{Datasets}
 \textbf{The CP2V$^2$ Dataset}: We have collected a dataset for the purpose of researching Camera Pose to Virtual Video reconstruction and we call this dataset CP2V$^2$. The data were obtained from two office environments using a robotic arm with a fixed base and an RGBD camera attached to it. CP2V$^2$ contains 22 $1024\times742$ videos taken at 20 frames per second while the camera's spatial position and orientation accuracies are respectively $0.100mm$ and 0.209\textdegree. This dataset will be made publicly available and more detailed description of the dataset can be found in the supplementary materials. 

 \textbf{The 7 Scene Dataset}: In addition to our own datasets, this work also conducted experiments on the publicly available 7scenes dataset \cite{r73}, which includes RGBD sample images with a resolution of 640×480 pixels in seven different indoor scenes. 

\begin{figure}[!ht]
\label{TrainingDataSplit}
\centering 
\subfigure[]{
\label{Fig.sub.1}
\includegraphics[scale=0.1830]{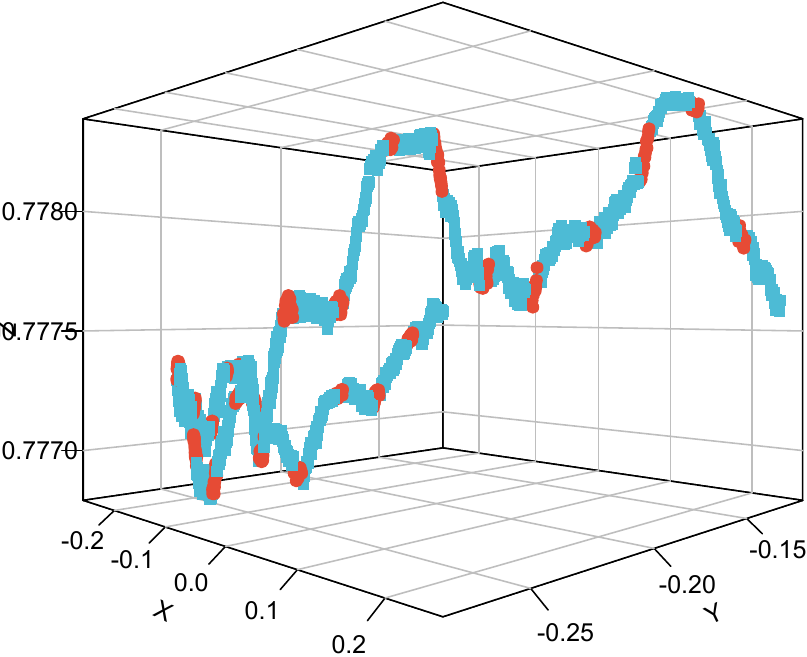}}
\subfigure[]{
\label{Fig.sub.2}
\includegraphics[scale=0.1830]{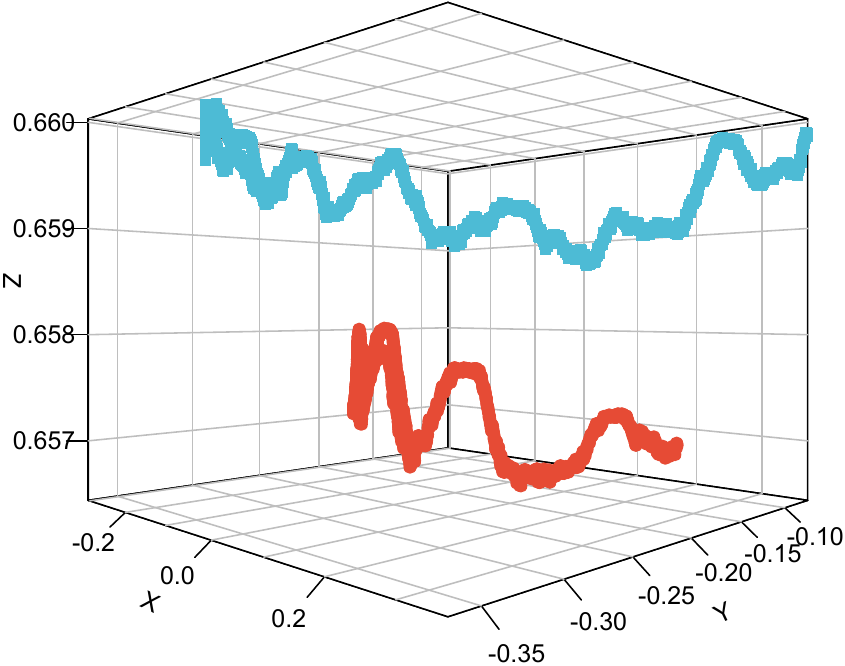}}
\subfigure[]{
\label{Fig.sub.3}
\includegraphics[scale=0.183]{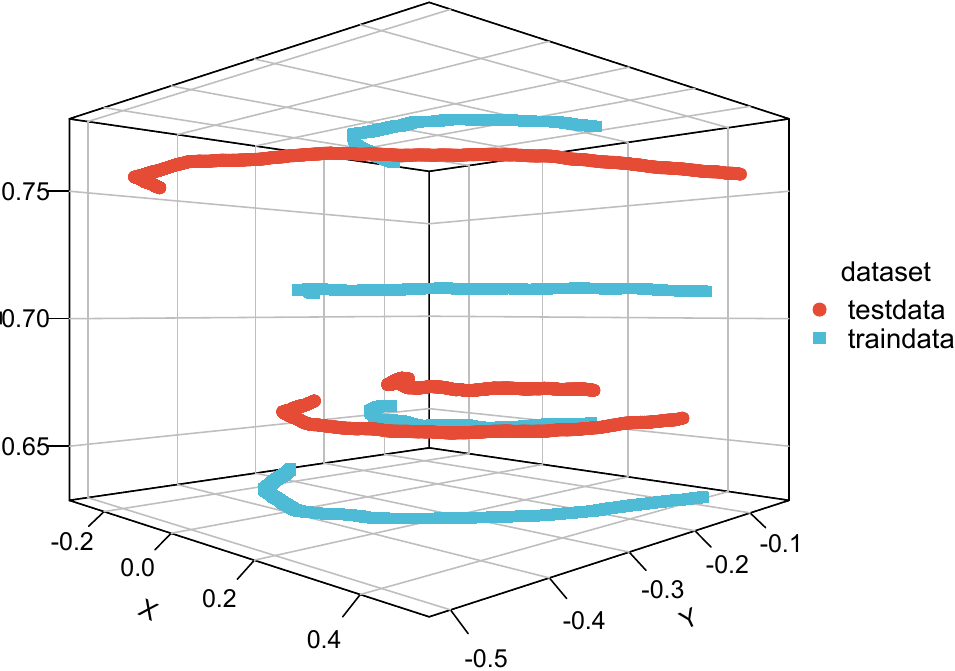}}
\caption{Illustration of different new view synthesis settings.}
\label{TrainingDataSplit1}
\end{figure}

\subsection{Results}

\begin{figure*}[!ht]
\label{TrainingDataSplit}
\centering 
\subfigure[PSNR]{
\label{Fig.sub.1}
\includegraphics[width=0.32\linewidth,height=2.0cm]{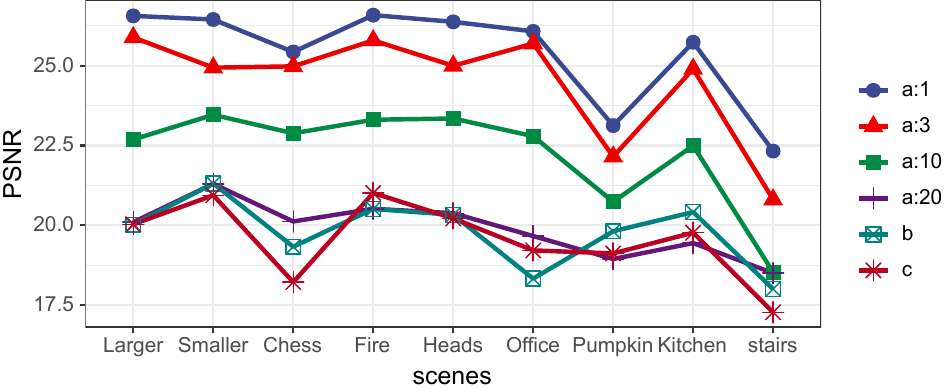}}
\subfigure[SSMI]{
\label{Fig.sub.2}
\includegraphics[width=0.32\linewidth,height=2.0cm]{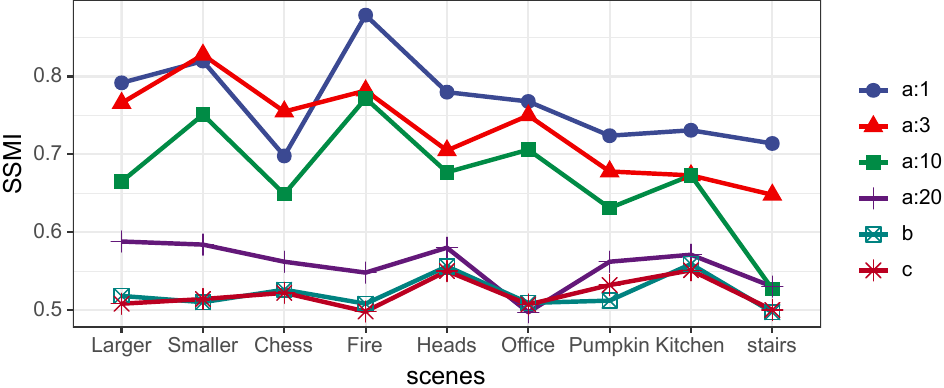}}
\subfigure[LPIPS]{
\label{Fig.sub.3}
\includegraphics[width=0.32\linewidth,height=2.0cm]{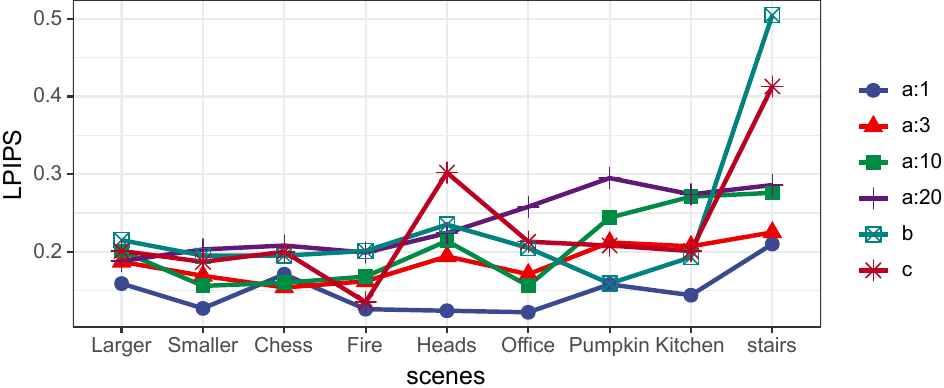}}
\caption{New view synthesis performances of P2I-NET in different experimental settings. These quantitative results are based on comparing the synthesised frames with their ground truth. a:x is the experimental setting of Fig. \ref{TrainingDataSplit1} (a), where x=1,3,10.20 is the number of the missing frames; b is the experimental setting of Fig.\ref{TrainingDataSplit1}(b), c is the experimental setting of Fig.\ref{TrainingDataSplit1}(c).}
\label{quantity result}
\end{figure*}

\textbf{New frame synthesis from frames on the same trajectory}: The first experiment tests P2I-NET's capability of synthesising missing frames from the same video sequence. We assume the $(100m + 1)^{th}, (100m + 2)^{th}, ..., (100m + N)^{th}$ frames are missing, where $m = 1, 2, ..., M$ and $(100M + N) \leq $ Total Number of Frames. Our task is to synthesis the $N\in \{1, 3, 10, 20\}$ missing frames. That is, for every 100 frames, we assume their subsequent 1, 3, 10 or 20 consecutive frames are missing. In the dataset, there are real frames in these locations which allowed us to measure the synthesis performances. Fig.\ref{TrainingDataSplit1}) (a) illustrates this scenario where the red-colored positions on the trajectory indicate missing frames. We use the rest of the frames from the same trajectory to train the P2I-NET.

\textbf{Single sequence video synthesis from a single adjacent trajectory}: In this experiment, we use frames from the entire sequence of one trajectory as training data to train a P2I-NET, and then test it on the entire sequence of a video taken from an adjacent trajectory. As is shown on Fig.\ref{TrainingDataSplit1} (b). 
It is worth noting that the dataset of training trajectory should include all the scene content of neighboring trajectory images.

\textbf{Multiple sequences synthesis from multiple adjacent trajectories}: In this experiment, we use frames from multiple trajectories as training data to train a P2I-NET, and then use the model to synthesis multiple sequences of videos on trajectories outside those in the training set. As is shown on Fig.\ref{TrainingDataSplit1} (c) .

\begin{figure*}[!htbp]
\label{QualitativeResults}
 \centering
\includegraphics[width=\textwidth,height=10cm]{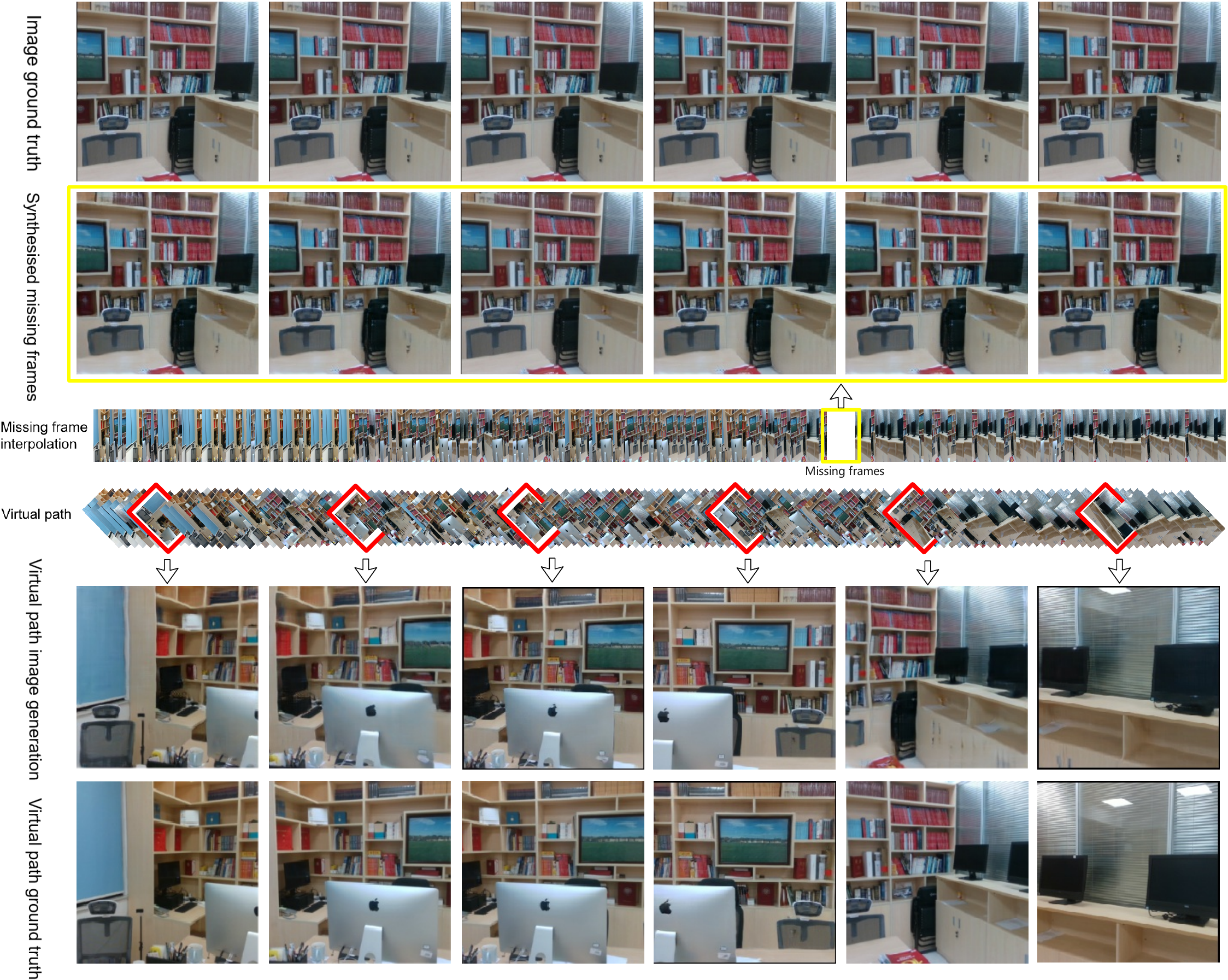}
  \caption{RGB image synthesis in larger office in dataset CP2V$^2$.The synthetic images in second row is missing frame from same trajectory , and the third row is the synthetic image from adjacent virtual camera trajectory . The fourth row is image ground truth from  adjacent trajectory .}
 \label{visual qualitative result in office}
\end{figure*}

These training and testing settings allowed us to evaluate the performance of our approach on different application scenarios, such as 
frame interpolation for slow motion videos, or generating virtual videos for different trajectories. Quantitative results of the above experiments are shown in Fig\ref{quantity result} and visual qualitative result examples are shown in Fig. \ref{visual qualitative result in office} As expected, with more missing frames, the synthesised image quality decreases. It is seen that for up to 3 missing frames, P2I-NET can predict the frames fairly well. It is also seen that P2I-NET can synthesis an entire trajectory of videos from models trained entirely outside of the current trajectory. These results demonstrate that the P2I-NET model has established intrinsic correspondence between pose and image content near the training data trajectory and can synthesise high-quality RGB images for new (virtual) camera viewpoints. 

\begin{table*}[!htbp]
\begin{center}
\begin{tabular}{c|c|cccccc}
\hline
 \multicolumn{2}{c|}{Methods}&Mip-NeRF\cite{barron2021mip} & Nerf \cite{r1}&  instant-ngp \cite{T2022Instant} &nerfacto \cite{2021NeRFactor} &volinga ai\cite{nerfstudio} &P2I-NET\\
 \hline
 

 \multirow{4}{7em}{ The CP2V$^2$ Dataset  (512×512 pixels)} & PSNR$\uparrow$ & 13.580 & 15.181 &16.449 & 17.510&16.242 & \textbf{20.898}\\
 & SSIM $\uparrow$ &0.436 & 0.397 &0.601 & 0.587&0.545 &\textbf{0.672}\\
 & LPIPS $\downarrow$ & 0.875 & 0.580 &0.573 & 0.461&0.420 & \textbf{0.197}\\

 & FPS $\uparrow$ & 0.0098 & 0.092 &0.139 & 0.730&0.857 & \textbf{100.653}\\
\hline
 \multirow{4}{5em}{ 7 scenes  \cite{r73}(256×256 pixels)} & PSNR$\uparrow$ & 15.698 & 12.524 &17.401 & \textbf{24.489}&21.835 & 21.801\\
 & SSIM $\uparrow$ & 0.618 & 0.333 &0.369 &\textbf{ 0.778}&0.676 & 0.753\\
 & LPIPS $\downarrow$  & 0.543 & 0.818 &0.772 & 0.159&0.194 & \textbf{0.149}\\

 & FPS $\uparrow$ & 0.031 & 0.563 &0.383 & 2.548&0.142 & \textbf{103.312}\\
\hline
\end{tabular}
\end{center}
\caption{Quantitatively comparison with top performing techniques in the literature. FPS is frames per second. }
\label{quantity}
\end{table*}

\begin{figure*}[!htbp]
 \centering

 \rotatebox{90}{\centering\hspace{0.cm} The CP2V$^2$ Dataset }
 {\includegraphics[width=0.1600\linewidth]{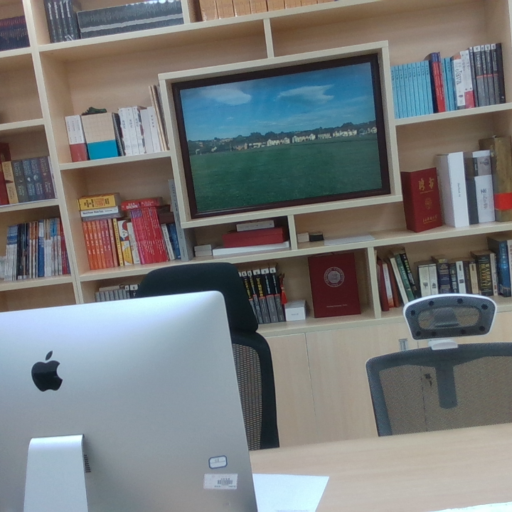}}
 {\includegraphics[width=0.1600\linewidth]{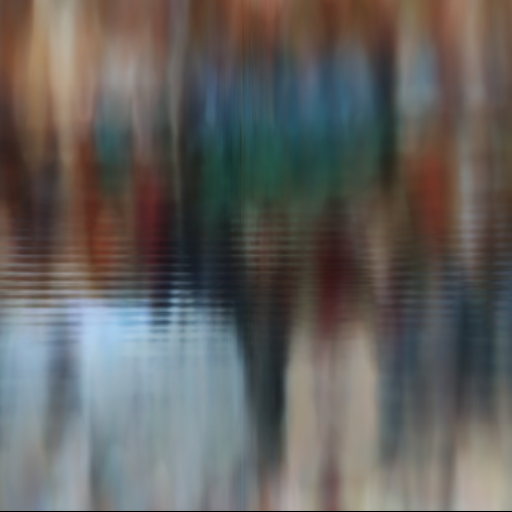}}
 {\includegraphics[width=0.160\linewidth]{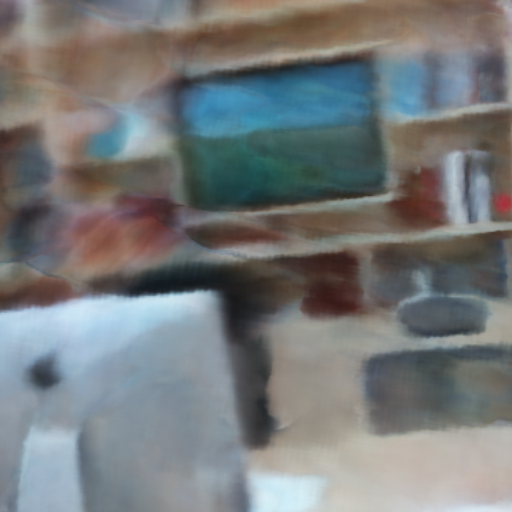}}
 {\includegraphics[width=0.160\linewidth]{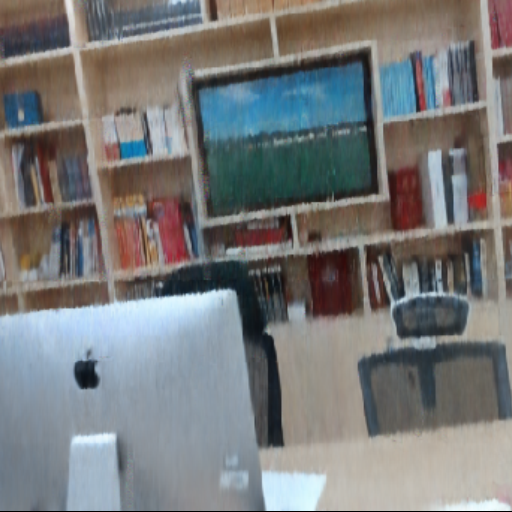}}
 {\includegraphics[width=0.160\linewidth]{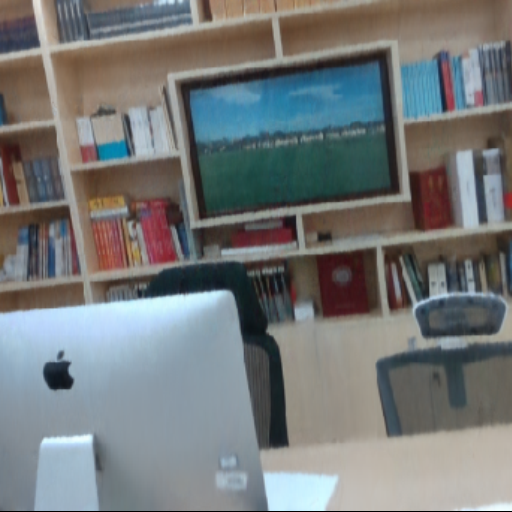}}
 {\includegraphics[width=0.160\linewidth]{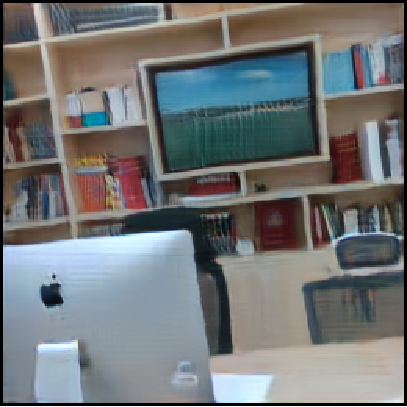}}

\rotatebox{90}{\centering\hspace{0.2cm} Fire in 7scenes\cite{r73}}
\subfigure[ground truth]{\includegraphics[width=0.16\linewidth]{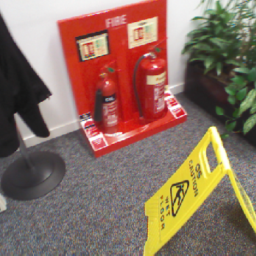}}
 \subfigure[Mip-NeRF]{\includegraphics[width=0.16\linewidth]{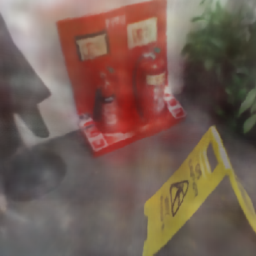}}
 \subfigure[NeRF]{\includegraphics[width=0.16\linewidth]{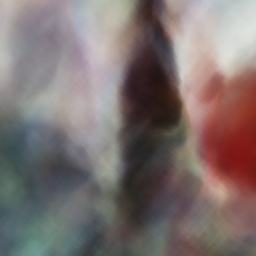}}
 \subfigure[Instant-ngp]{\includegraphics[width=0.16\linewidth]{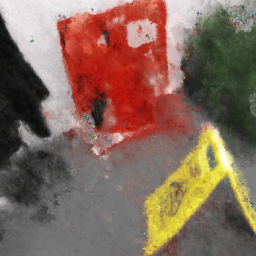}}
 \subfigure[NeRFacto]{\includegraphics[width=0.16\linewidth]{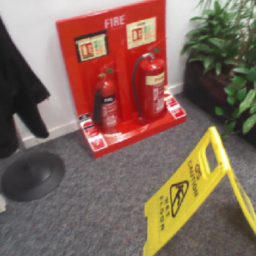}}
\subfigure[P2I-NET]{\includegraphics[width=0.16\linewidth]{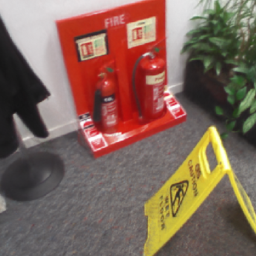}}
\caption{Visual comparison of new view synthesis results with top performing techniques in the literature at the setting in Figure 2(b) as the baseline , It is seen that P2I-NET has a better capability in representing finer details of the geometry and texture across the generated views compared to most other neural radiance fields (NERF) methods.}
 \label{generalization comparison}
\end{figure*}

\begin{table*}[!]
\begin{center}
\begin{tabular}{c|ccccc|ccc}
\hline
 &Input &Attention block & Feature channel(G)& Pose consistency & EnNet & PSNR$\uparrow$ & SSIM$\uparrow$ & LPIPS $\downarrow$\\ 
\hline
(1) Minimalist version & RGB & excludeing &fewer&None &  excludeing& 9.492 & 0.209 & 0.626 \\
(2) No depth data & RGB &including&more & HD,LD & excludeing & 19.696 & 0.721 & 0.170 \\
(3) No pose matching in HD & RGBD  &including&more& LD & excludeing & 11.178& 0.502 & 0.525\\ 
(4) No pose matching in LD & RGBD &including&more& HD  & excludeing & 18.982 & 0.535 & 0.281 \\
(5) No attention block & RGBD &  excludeing&more &HD,LD& excludeing& 18.307 & 0.585 & 0.342 \\
(6) \textbf{Fewer} feature channel & RGBD &  includeing&fewer &HD,LD& excludeing& 18.793 & 0.632 & 0.314 \\
(7) No image enhancement & RGBD &including&more& HD,LD &  excludeing & 19.781& 0.752 & 0.157 \\   
(8) Completed model & RGBD&including&more & HD, LD & including& 21.190 & 0.763 & 0.121 \\
 \hline
\end{tabular}
\end{center}
\caption{An ablation study of our model. The results are averages over the larger and smaller office scenes data, \textbf{HD} stands for pose consistency constrain in the latent space and \textbf{LD} in real world pose space.}
\label{table:ablation}
\end{table*}

\begin{figure*}[!]
 \centering
 \subfigure[groundtruth]{{\includegraphics[width=0.161\linewidth]{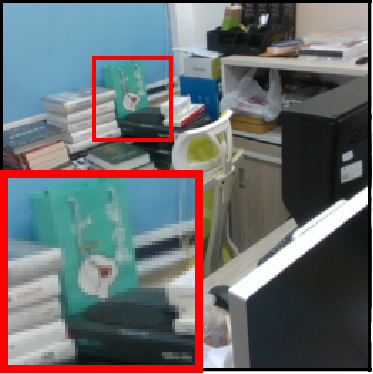}}}
 \subfigure[P2I-NET]{{\includegraphics[width=0.161\linewidth]{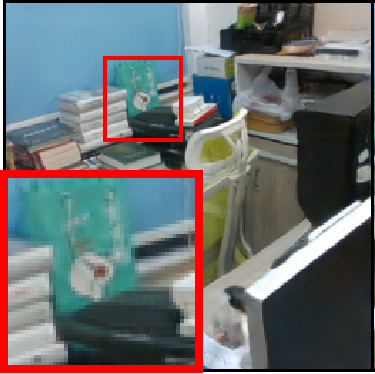}}}
 \subfigure[CNN]{{\includegraphics[width=0.161\linewidth]{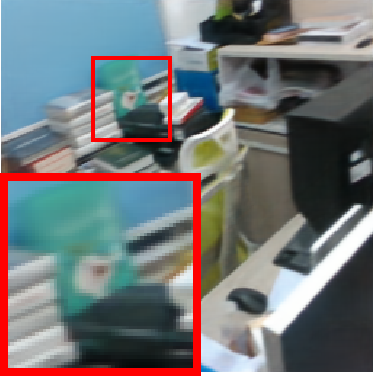}}}
 \subfigure[groundtruth]{{\includegraphics[width=0.161\linewidth]{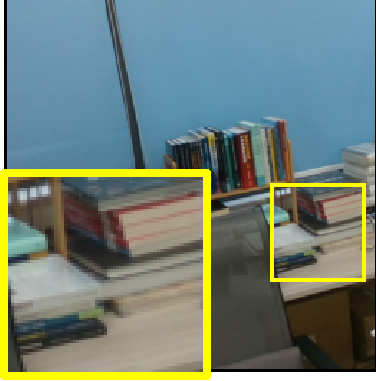}}}
 \subfigure[P2I-NET]{{\includegraphics[width=0.161\linewidth]{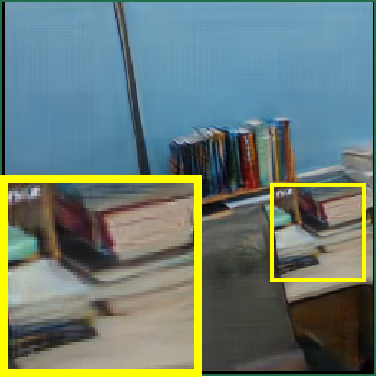}}}
 \subfigure[CNN]{{\includegraphics[width=0.161\linewidth]{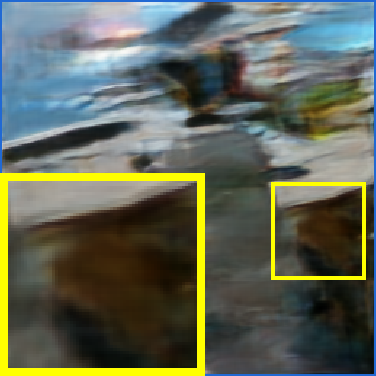}}}
\caption{ (b) (c) are test images generated from the same trajectory, (e) (f) are test images generated from the adjacent trajectory}
 \label{generalization comparison}
\end{figure*}

\subsection{Comparison}


For further evaluating our model, we compared it against current top-performing techniques for RGB view synthesis including {NeRF}\cite{r1}, {Mip-NeRF}\cite{barron2021mip}, {Nerfacto} \cite{2021NeRFactor}, volinga ai\cite{nerfstudio}, and {Instant-ngp} \cite{T2022Instant}. 
Quantitative and qualitative comparison of new view synthesis results with these methods are shown in Table \ref{quantity} and Fig. \ref{generalization comparison} respectively. 
These results clearly show that new view synthesis results of the P2I-NET are comparable or even superior to the results of other methods tested. Importantly, P2I-NET is computationally much more efficient than these methods. For synthesising larger image size ($512\times512$), P2I-NET is over \textbf{110} times faster than the next most efficient method (volinga ai\cite{nerfstudio}), while for smaller image size ($256\times 256$), P2I-NET is about \textbf{40} times faster than the next fastest method (nerfacto). The reason that P2I-NET has such computational advantage over other methods lies in its ability to directly establish the correspondence between the camera pose and the RGBD image without the need for accurate 3D scene construction from multiple viewpoints. This simplifies the image generation process, making it more efficient and faster. In contrast, other methods such as mip-nerf, nerf, instant-ngp, and nerfacto require the implicit construction of an accurate 3D model of the scene and use ray tracing to simulates the entire process of image generation, which limits their applicability in scenarios where on each location of the camera trajectory there is only one image from a single view point. Moreover, P2I-NET generates images end-to-end through CNNs, which results in fast image rendering, 
making it suitable for real-time applications where fast image generation is essential. 

\subsection{Ablation}
We have conducted an extensive ablation study to validate the design choices and parameters of our algorithm on our new CP2V$^2$ dataset.
Qualitative and intuitive explanations can be found in the supplementary material.Table \ref{table:ablation} show the results of different settings for the P2I-NET where row 8 is complete model serving as the reference. In row 1, we implemented a minimalist version of our model without depth data, pose matching in high-dimensional latent space (HD) and in Low-dimensional world pose space (LD), attention block , image enhancement network, and feature channels in every hidden layer of the generator are all set to 64 (denoted as "fewer" in Table \ref{table:ablation}). Note that in the complete model the number of feature channels in $G$ gradually increase from 64 (first layer) to 1024 (last layer). In rows 2-7, we removed these six components two at a time (excluding image enhancement for each, as it would disrupt the effectiveness of each component), and observed that each component provided quantitative benefits.

Results in rows 3-4 and 6 show that the model's performance decreases significantly without pose consistency constraint in both low-dimensional real world pose space and high-dimensional latent feature spaces. The reason is that if the pose consistency is only performed in a high-dimensional latent space or a low-dimensional world pose space, the matching of the pose and the image will not be very accurate, resulting in an overall pixel offset between the generated image and the real image. From the results in rows 2 and 7, we can infer that depth data provides extra geometric constraints, which allows the discriminator to better describe the pose conditional distribution $p(I|y)$ in the training data. Furthermore, by using both geometric and texture information to determine the pose, the discriminator $D$ can improve pose consistency in the world pose space. The results in rows 6 and 7 demonstrate that a fewer number of feature channels in the generator hidden layer cause the generated images to be less realistic and cannot match the ground truth in terms of color and object contour. The reason is that fewer feature channels can not store enough information to synthesising images with complex surface and texture details. 
The results in rows 5 and 7 show that discriminator $D$ with attention block focus on key feature in the images to better fit the pose conditional distribution $p(I|y)$ to the samples and obtain a higher precision mapping pose. 
The results in rows 7 and 8 demonstrate that there is no difference in the structure (reflected in the SSIM and LPIPS values) between the synthetic image and the ground truth without image enhancement. However, the image detail have a significant difference as reflected in the PSNR values. We can infer that the generator $G$ with a simple structure can not generate images with the same pixel detail as the ground truth.

\subsection{Discussion}
If a conventional CNN network is trained with the camera pose as input and the ground truth image as the supervisory signal, it may only fit the pose and the image rigidly rather than learning the pose conditional distribution that constructs the intrinsic correspondence between the pose and the image within the training scene. As shown in the left three images in Fig \ref{generalization comparison}, the interpolated images generated by the trained CNN network from the same trajectory appear blurred as the pixel values in the synthesis image are obtained by averaging the pixel values among neighboring frames. Moreover, as shown in the right three images in Fig \ref{generalization comparison}, when new viewpoint images in an adjacent trajectory are generated by the CNN network, they may lose their structure and content. The P2I-NET network has established an intrinsic correspondence between poses and images near the training data by learning the distribution $p(I|y)$. It has successfully generated images that are consistent with the ground truth in terms of content and structure. The more discreet sample data can better represent the scene space distribution $p(I|y)$, resulting in more realistic and accurate images synthesis. Finally, the trained generator in the P2I-NET network will automatically creates a virtual camera nearby the training scene.

\section{Conclusion}
In this study, we constructed the P2I-NET network that employs adversarial training to capture image pose conditional distribution, establishing an intrinsic correspondence between pose and image for new view image synthesis. we implement experiment on the CP2V$^2$ dataset and public 7 scenes dataset.The experiment resuts demonstrates that the trained P2I-NET network can perform high-quality image interpolation on the same trajectory, and even generate RGB images for adjacent virtual camera trajectory.This method has wide-ranging applications, including the creation of synthetic training data for visual recognition and providing additional sample image data for SLAM and SFM. Moreover, the efficiency of the network enables it to be used in real-time applications such as virtual reality and augmented reality. Overall, the P2I-NET network presents a promising framework for efficient and accurate image synthesis with potential applications in various fields of computer vision.




\begin{acks}
This work was supported in part by the National Natural Science Foundation of China under grants U22B2035 and 62271323, Guangdong Basic and Applied Basic Research Foundation under grant 2021A1515011584 and 2023A1515012956, Shenzhen R\&D Program under grants JCYJ20220531102408020 and JCYJ20200109105008228, and in part by Shenzhen Institute for Artificial Intelligence and Robotics for Society (AIRS) and Guangdong Key Laboratory of Intelligent Information Processing.
\end{acks}

\bibliographystyle{ACM-Reference-Format}
\balance
\bibliography{sample-base}


\begin{thebibliography}{35}


\ifx \showCODEN    \undefined \def \showCODEN     #1{\unskip}     \fi
\ifx \showDOI      \undefined \def \showDOI       #1{#1}\fi
\ifx \showISBNx    \undefined \def \showISBNx     #1{\unskip}     \fi
\ifx \showISBNxiii \undefined \def \showISBNxiii  #1{\unskip}     \fi
\ifx \showISSN     \undefined \def \showISSN      #1{\unskip}     \fi
\ifx \showLCCN     \undefined \def \showLCCN      #1{\unskip}     \fi
\ifx \shownote     \undefined \def \shownote      #1{#1}          \fi
\ifx \showarticletitle \undefined \def \showarticletitle #1{#1}   \fi
\ifx \showURL      \undefined \def \showURL       {\relax}        \fi
\providecommand\bibfield[2]{#2}
\providecommand\bibinfo[2]{#2}
\providecommand\natexlab[1]{#1}
\providecommand\showeprint[2][]{arXiv:#2}

\bibitem[201(2018)]%
        {2018Neural}
 \bibinfo{year}{2018}\natexlab{}.
\newblock \showarticletitle{Neural scene representation and rendering}.
\newblock \bibinfo{journal}{\emph{Science}} \bibinfo{volume}{360}, \bibinfo{number}{6394} (\bibinfo{year}{2018}), \bibinfo{pages}{1204--1210}.
\newblock


\bibitem[Abu~Alhaija et~al\mbox{.}(2019)]%
        {r92}
\bibfield{author}{\bibinfo{person}{Hassan Abu~Alhaija}, \bibinfo{person}{Siva~Karthik Mustikovela}, \bibinfo{person}{Andreas Geiger}, {and} \bibinfo{person}{Carsten Rother}.} \bibinfo{year}{2019}\natexlab{}.
\newblock \showarticletitle{Geometric image synthesis}. In \bibinfo{booktitle}{\emph{Computer Vision--ACCV 2018: 14th Asian Conference on Computer Vision, Perth, Australia, December 2--6, 2018, Revised Selected Papers, Part VI 14}}. Springer, \bibinfo{pages}{85--100}.
\newblock


\bibitem[Barron et~al\mbox{.}(2021)]%
        {barron2021mip}
\bibfield{author}{\bibinfo{person}{Jonathan~T Barron}, \bibinfo{person}{Ben Mildenhall}, \bibinfo{person}{Matthew Tancik}, \bibinfo{person}{Peter Hedman}, \bibinfo{person}{Ricardo Martin-Brualla}, {and} \bibinfo{person}{Pratul~P Srinivasan}.} \bibinfo{year}{2021}\natexlab{}.
\newblock \showarticletitle{Mip-nerf: A multiscale representation for anti-aliasing neural radiance fields}. In \bibinfo{booktitle}{\emph{Proceedings of the IEEE/CVF International Conference on Computer Vision}}. \bibinfo{pages}{5855--5864}.
\newblock


\bibitem[Chabra et~al\mbox{.}(2020)]%
        {r91}
\bibfield{author}{\bibinfo{person}{Rohan Chabra}, \bibinfo{person}{Jan~E Lenssen}, \bibinfo{person}{Eddy Ilg}, \bibinfo{person}{Tanner Schmidt}, \bibinfo{person}{Julian Straub}, \bibinfo{person}{Steven Lovegrove}, {and} \bibinfo{person}{Richard Newcombe}.} \bibinfo{year}{2020}\natexlab{}.
\newblock \showarticletitle{Deep local shapes: Learning local sdf priors for detailed 3d reconstruction}. In \bibinfo{booktitle}{\emph{Computer Vision--ECCV 2020: 16th European Conference, Glasgow, UK, August 23--28, 2020, Proceedings, Part XXIX 16}}. Springer, \bibinfo{pages}{608--625}.
\newblock


\bibitem[Curless and Levoy(1996)]%
        {r85}
\bibfield{author}{\bibinfo{person}{Brian Curless} {and} \bibinfo{person}{Marc Levoy}.} \bibinfo{year}{1996}\natexlab{}.
\newblock \showarticletitle{A volumetric method for building complex models from range images}. In \bibinfo{booktitle}{\emph{Proceedings of the 23rd annual conference on Computer graphics and interactive techniques}}. \bibinfo{pages}{303--312}.
\newblock


\bibitem[Genova et~al\mbox{.}(2020)]%
        {r102}
\bibfield{author}{\bibinfo{person}{Kyle Genova}, \bibinfo{person}{Forrester Cole}, \bibinfo{person}{Avneesh Sud}, \bibinfo{person}{Aaron Sarna}, {and} \bibinfo{person}{Thomas Funkhouser}.} \bibinfo{year}{2020}\natexlab{}.
\newblock \showarticletitle{Local deep implicit functions for 3d shape}. In \bibinfo{booktitle}{\emph{Proceedings of the IEEE/CVF Conference on Computer Vision and Pattern Recognition}}. \bibinfo{pages}{4857--4866}.
\newblock


\bibitem[Gonzalez(2002)]%
        {2002Woods}
\bibfield{author}{\bibinfo{person}{R. Gonzalez}.} \bibinfo{year}{2002}\natexlab{}.
\newblock \showarticletitle{Woods RE: Digital Image Processing}.
\newblock \bibinfo{journal}{\emph{upper saddle river nj pearson/prentice hall}} (\bibinfo{year}{2002}).
\newblock


\bibitem[Goodfellow et~al\mbox{.}(2020)]%
        {goodfellow2020generative}
\bibfield{author}{\bibinfo{person}{Ian Goodfellow}, \bibinfo{person}{Jean Pouget-Abadie}, \bibinfo{person}{Mehdi Mirza}, \bibinfo{person}{Bing Xu}, \bibinfo{person}{David Warde-Farley}, \bibinfo{person}{Sherjil Ozair}, \bibinfo{person}{Aaron Courville}, {and} \bibinfo{person}{Yoshua Bengio}.} \bibinfo{year}{2020}\natexlab{}.
\newblock \showarticletitle{Generative adversarial networks}.
\newblock \bibinfo{journal}{\emph{Commun. ACM}} \bibinfo{volume}{63}, \bibinfo{number}{11} (\bibinfo{year}{2020}), \bibinfo{pages}{139--144}.
\newblock


\bibitem[He et~al\mbox{.}(2016)]%
        {r4}
\bibfield{author}{\bibinfo{person}{Kaiming He}, \bibinfo{person}{Xiangyu Zhang}, \bibinfo{person}{Shaoqing Ren}, {and} \bibinfo{person}{Jian Sun}.} \bibinfo{year}{2016}\natexlab{}.
\newblock \showarticletitle{Deep residual learning for image recognition}. In \bibinfo{booktitle}{\emph{Proceedings of the IEEE conference on computer vision and pattern recognition}}. \bibinfo{pages}{770--778}.
\newblock


\bibitem[Henzler et~al\mbox{.}(2020)]%
        {r86}
\bibfield{author}{\bibinfo{person}{Philipp Henzler}, \bibinfo{person}{Niloy~J Mitra}, {and} \bibinfo{person}{Tobias Ritschel}.} \bibinfo{year}{2020}\natexlab{}.
\newblock \showarticletitle{Learning a neural 3d texture space from 2d exemplars}. In \bibinfo{booktitle}{\emph{Proceedings of the IEEE/CVF Conference on Computer Vision and Pattern Recognition}}. \bibinfo{pages}{8356--8364}.
\newblock


\bibitem[Liu et~al\mbox{.}(2020)]%
        {liu2020disentangled}
\bibfield{author}{\bibinfo{person}{Xiaoming Liu}, \bibinfo{person}{Luan~Quoc Tran}, {and} \bibinfo{person}{Xi Yin}.} \bibinfo{year}{2020}\natexlab{}.
\newblock \bibinfo{title}{Disentangled representation learning generative adversarial network for pose-invariant face recognition}.
\newblock
\newblock
\newblock
\shownote{US Patent App. 16/648,202}.


\bibitem[Mescheder et~al\mbox{.}(2019)]%
        {r103}
\bibfield{author}{\bibinfo{person}{Lars Mescheder}, \bibinfo{person}{Michael Oechsle}, \bibinfo{person}{Michael Niemeyer}, \bibinfo{person}{Sebastian Nowozin}, {and} \bibinfo{person}{Andreas Geiger}.} \bibinfo{year}{2019}\natexlab{}.
\newblock \showarticletitle{Occupancy networks: Learning 3d reconstruction in function space}. In \bibinfo{booktitle}{\emph{Proceedings of the IEEE/CVF conference on computer vision and pattern recognition}}. \bibinfo{pages}{4460--4470}.
\newblock


\bibitem[Mildenhall et~al\mbox{.}(2021)]%
        {r1}
\bibfield{author}{\bibinfo{person}{Ben Mildenhall}, \bibinfo{person}{Pratul~P Srinivasan}, \bibinfo{person}{Matthew Tancik}, \bibinfo{person}{Jonathan~T Barron}, \bibinfo{person}{Ravi Ramamoorthi}, {and} \bibinfo{person}{Ren Ng}.} \bibinfo{year}{2021}\natexlab{}.
\newblock \showarticletitle{Nerf: Representing scenes as neural radiance fields for view synthesis}.
\newblock \bibinfo{journal}{\emph{Commun. ACM}} \bibinfo{volume}{65}, \bibinfo{number}{1} (\bibinfo{year}{2021}), \bibinfo{pages}{99--106}.
\newblock


\bibitem[Miyato et~al\mbox{.}(2018)]%
        {r71}
\bibfield{author}{\bibinfo{person}{Takeru Miyato}, \bibinfo{person}{Toshiki Kataoka}, \bibinfo{person}{Masanori Koyama}, {and} \bibinfo{person}{Yuichi Yoshida}.} \bibinfo{year}{2018}\natexlab{}.
\newblock \showarticletitle{Spectral normalization for generative adversarial networks}.
\newblock \bibinfo{journal}{\emph{arXiv preprint arXiv:1802.05957}} (\bibinfo{year}{2018}).
\newblock


\bibitem[M{\"u}ller et~al\mbox{.}(2022)]%
        {muller2022instant}
\bibfield{author}{\bibinfo{person}{Thomas M{\"u}ller}, \bibinfo{person}{Alex Evans}, \bibinfo{person}{Christoph Schied}, \bibinfo{person}{Marco Foco}, \bibinfo{person}{Andr{\'a}s B{\'o}dis-Szomor{\'u}}, \bibinfo{person}{Isaac Deutsch}, \bibinfo{person}{Michael Shelley}, {and} \bibinfo{person}{Alexander Keller}.} \bibinfo{year}{2022}\natexlab{}.
\newblock \showarticletitle{Instant neural radiance fields}.
\newblock In \bibinfo{booktitle}{\emph{ACM SIGGRAPH 2022 Real-Time Live!}} \bibinfo{pages}{1--2}.
\newblock


\bibitem[Müller et~al\mbox{.}(2022)]%
        {T2022Instant}
\bibfield{author}{\bibinfo{person}{T Müller}, \bibinfo{person}{A. Evans}, \bibinfo{person}{C. Schied}, {and} \bibinfo{person}{A. Keller}.} \bibinfo{year}{2022}\natexlab{}.
\newblock \showarticletitle{Instant Neural Graphics Primitives with a Multiresolution Hash Encoding}.
\newblock \bibinfo{journal}{\emph{arXiv e-prints}} (\bibinfo{year}{2022}).
\newblock


\bibitem[Nguyen-Phuoc et~al\mbox{.}(2019)]%
        {r93}
\bibfield{author}{\bibinfo{person}{Thu Nguyen-Phuoc}, \bibinfo{person}{Chuan Li}, \bibinfo{person}{Lucas Theis}, \bibinfo{person}{Christian Richardt}, {and} \bibinfo{person}{Yong-Liang Yang}.} \bibinfo{year}{2019}\natexlab{}.
\newblock \showarticletitle{Hologan: Unsupervised learning of 3d representations from natural images}. In \bibinfo{booktitle}{\emph{Proceedings of the IEEE/CVF International Conference on Computer Vision}}. \bibinfo{pages}{7588--7597}.
\newblock


\bibitem[Noguchi and Harada(2019)]%
        {2019RGBD}
\bibfield{author}{\bibinfo{person}{A. Noguchi} {and} \bibinfo{person}{T. Harada}.} \bibinfo{year}{2019}\natexlab{}.
\newblock \bibinfo{title}{RGBD-GAN: Unsupervised 3D Representation Learning From Natural Image Datasets via RGBD Image Synthesis}.
\newblock
\newblock


\bibitem[Novotny et~al\mbox{.}(2019)]%
        {novotny2019perspectivenet}
\bibfield{author}{\bibinfo{person}{David Novotny}, \bibinfo{person}{Ben Graham}, {and} \bibinfo{person}{Jeremy Reizenstein}.} \bibinfo{year}{2019}\natexlab{}.
\newblock \showarticletitle{Perspectivenet: A scene-consistent image generator for new view synthesis in real indoor environments}.
\newblock \bibinfo{journal}{\emph{Advances in Neural Information Processing Systems}}  \bibinfo{volume}{32} (\bibinfo{year}{2019}).
\newblock


\bibitem[Oechsle et~al\mbox{.}(2019a)]%
        {r87}
\bibfield{author}{\bibinfo{person}{Michael Oechsle}, \bibinfo{person}{Lars Mescheder}, \bibinfo{person}{Michael Niemeyer}, \bibinfo{person}{Thilo Strauss}, {and} \bibinfo{person}{Andreas Geiger}.} \bibinfo{year}{2019}\natexlab{a}.
\newblock \showarticletitle{Texture fields: Learning texture representations in function space}. In \bibinfo{booktitle}{\emph{Proceedings of the IEEE/CVF International Conference on Computer Vision}}. \bibinfo{pages}{4531--4540}.
\newblock


\bibitem[Oechsle et~al\mbox{.}(2019b)]%
        {r100}
\bibfield{author}{\bibinfo{person}{Michael Oechsle}, \bibinfo{person}{Lars Mescheder}, \bibinfo{person}{Michael Niemeyer}, \bibinfo{person}{Thilo Strauss}, {and} \bibinfo{person}{Andreas Geiger}.} \bibinfo{year}{2019}\natexlab{b}.
\newblock \showarticletitle{Texture fields: Learning texture representations in function space}. In \bibinfo{booktitle}{\emph{Proceedings of the IEEE/CVF International Conference on Computer Vision}}. \bibinfo{pages}{4531--4540}.
\newblock


\bibitem[Rainer et~al\mbox{.}(2020)]%
        {r88}
\bibfield{author}{\bibinfo{person}{Gilles Rainer}, \bibinfo{person}{Abhijeet Ghosh}, \bibinfo{person}{Wenzel Jakob}, {and} \bibinfo{person}{Tim Weyrich}.} \bibinfo{year}{2020}\natexlab{}.
\newblock \showarticletitle{Unified neural encoding of BTFs}. In \bibinfo{booktitle}{\emph{Computer Graphics Forum}}, Vol.~\bibinfo{volume}{39}. Wiley Online Library, \bibinfo{pages}{167--178}.
\newblock


\bibitem[Rainer et~al\mbox{.}(2019)]%
        {r89}
\bibfield{author}{\bibinfo{person}{Gilles Rainer}, \bibinfo{person}{Wenzel Jakob}, \bibinfo{person}{Abhijeet Ghosh}, {and} \bibinfo{person}{Tim Weyrich}.} \bibinfo{year}{2019}\natexlab{}.
\newblock \showarticletitle{Neural BTF compression and interpolation}. In \bibinfo{booktitle}{\emph{Computer Graphics Forum}}, Vol.~\bibinfo{volume}{38}. Wiley Online Library, \bibinfo{pages}{235--244}.
\newblock


\bibitem[Ren et~al\mbox{.}(2013)]%
        {r90}
\bibfield{author}{\bibinfo{person}{Peiran Ren}, \bibinfo{person}{Jiaping Wang}, \bibinfo{person}{Minmin Gong}, \bibinfo{person}{Stephen Lin}, \bibinfo{person}{Xin Tong}, {and} \bibinfo{person}{Baining Guo}.} \bibinfo{year}{2013}\natexlab{}.
\newblock \showarticletitle{Global illumination with radiance regression functions.}
\newblock \bibinfo{journal}{\emph{ACM Trans. Graph.}} \bibinfo{volume}{32}, \bibinfo{number}{4} (\bibinfo{year}{2013}), \bibinfo{pages}{130--1}.
\newblock


\bibitem[Rombach et~al\mbox{.}(2022)]%
        {rombach2022high}
\bibfield{author}{\bibinfo{person}{Robin Rombach}, \bibinfo{person}{Andreas Blattmann}, \bibinfo{person}{Dominik Lorenz}, \bibinfo{person}{Patrick Esser}, {and} \bibinfo{person}{Bj{\"o}rn Ommer}.} \bibinfo{year}{2022}\natexlab{}.
\newblock \showarticletitle{High-resolution image synthesis with latent diffusion models}. In \bibinfo{booktitle}{\emph{Proceedings of the IEEE/CVF Conference on Computer Vision and Pattern Recognition}}. \bibinfo{pages}{10684--10695}.
\newblock


\bibitem[Shen et~al\mbox{.}(2018)]%
        {2018FaceID}
\bibfield{author}{\bibinfo{person}{Y. Shen}, \bibinfo{person}{P. Luo}, \bibinfo{person}{J. Yan}, \bibinfo{person}{X. Wang}, {and} \bibinfo{person}{X. Tang}.} \bibinfo{year}{2018}\natexlab{}.
\newblock \showarticletitle{FaceID-GAN: Learning a Symmetry Three-Player GAN for Identity-Preserving Face Synthesis}.
\newblock \bibinfo{journal}{\emph{IEEE}} (\bibinfo{year}{2018}).
\newblock


\bibitem[Shotton et~al\mbox{.}(2013)]%
        {r73}
\bibfield{author}{\bibinfo{person}{Jamie Shotton}, \bibinfo{person}{Ben Glocker}, \bibinfo{person}{Christopher Zach}, \bibinfo{person}{Shahram Izadi}, \bibinfo{person}{Antonio Criminisi}, {and} \bibinfo{person}{Andrew Fitzgibbon}.} \bibinfo{year}{2013}\natexlab{}.
\newblock \showarticletitle{Scene coordinate regression forests for camera relocalization in RGB-D images}. In \bibinfo{booktitle}{\emph{Proceedings of the IEEE conference on computer vision and pattern recognition}}. \bibinfo{pages}{2930--2937}.
\newblock


\bibitem[Simonyan and Zisserman(2014)]%
        {r3}
\bibfield{author}{\bibinfo{person}{Karen Simonyan} {and} \bibinfo{person}{Andrew Zisserman}.} \bibinfo{year}{2014}\natexlab{}.
\newblock \showarticletitle{Very deep convolutional networks for large-scale image recognition}.
\newblock \bibinfo{journal}{\emph{arXiv preprint arXiv:1409.1556}} (\bibinfo{year}{2014}).
\newblock


\bibitem[Sitzmann et~al\mbox{.}(2018)]%
        {2018DeepVoxels}
\bibfield{author}{\bibinfo{person}{V. Sitzmann}, \bibinfo{person}{J. Thies}, \bibinfo{person}{F. Heide}, \bibinfo{person}{M Nießner}, \bibinfo{person}{G. Wetzstein}, {and} \bibinfo{person}{M Zollhöfer}.} \bibinfo{year}{2018}\natexlab{}.
\newblock \showarticletitle{DeepVoxels: Learning Persistent 3D Feature Embeddings}.
\newblock  (\bibinfo{year}{2018}).
\newblock


\bibitem[Tancik et~al\mbox{.}(2023)]%
        {nerfstudio}
\bibfield{author}{\bibinfo{person}{Matthew Tancik}, \bibinfo{person}{Ethan Weber}, \bibinfo{person}{Evonne Ng}, \bibinfo{person}{Ruilong Li}, \bibinfo{person}{Brent Yi}, \bibinfo{person}{Justin Kerr}, \bibinfo{person}{Terrance Wang}, \bibinfo{person}{Alexander Kristoffersen}, \bibinfo{person}{Jake Austin}, \bibinfo{person}{Kamyar Salahi}, \bibinfo{person}{Abhik Ahuja}, \bibinfo{person}{David McAllister}, {and} \bibinfo{person}{Angjoo Kanazawa}.} \bibinfo{year}{2023}\natexlab{}.
\newblock \showarticletitle{Nerfstudio: A Modular Framework for Neural Radiance Field Development}.
\newblock \bibinfo{journal}{\emph{arXiv preprint arXiv:2302.04264}} (\bibinfo{year}{2023}).
\newblock


\bibitem[Wang and Gupta(2016)]%
        {r97}
\bibfield{author}{\bibinfo{person}{Xiaolong Wang} {and} \bibinfo{person}{Abhinav Gupta}.} \bibinfo{year}{2016}\natexlab{}.
\newblock \showarticletitle{Generative image modeling using style and structure adversarial networks}. In \bibinfo{booktitle}{\emph{Computer Vision--ECCV 2016: 14th European Conference, Amsterdam, The Netherlands, October 11--14, 2016, Proceedings, Part IV 14}}. Springer, \bibinfo{pages}{318--335}.
\newblock


\bibitem[Zhang and Agrawala(2023)]%
        {zhang2023adding}
\bibfield{author}{\bibinfo{person}{Lvmin Zhang} {and} \bibinfo{person}{Maneesh Agrawala}.} \bibinfo{year}{2023}\natexlab{}.
\newblock \showarticletitle{Adding conditional control to text-to-image diffusion models}.
\newblock \bibinfo{journal}{\emph{arXiv preprint arXiv:2302.05543}} (\bibinfo{year}{2023}).
\newblock


\bibitem[Zhang et~al\mbox{.}(2021)]%
        {2021NeRFactor}
\bibfield{author}{\bibinfo{person}{X. Zhang}, \bibinfo{person}{P.~P. Srinivasan}, \bibinfo{person}{B. Deng}, \bibinfo{person}{P. Debevec}, \bibinfo{person}{W.~T. Freeman}, {and} \bibinfo{person}{J.~T. Barron}.} \bibinfo{year}{2021}\natexlab{}.
\newblock \showarticletitle{NeRFactor: Neural Factorization of Shape and Reflectance Under an Unknown Illumination}.
\newblock  (\bibinfo{year}{2021}).
\newblock


\bibitem[Zhu et~al\mbox{.}(2018a)]%
        {r98}
\bibfield{author}{\bibinfo{person}{Jun-Yan Zhu}, \bibinfo{person}{Zhoutong Zhang}, \bibinfo{person}{Chengkai Zhang}, \bibinfo{person}{Jiajun Wu}, \bibinfo{person}{Antonio Torralba}, \bibinfo{person}{Josh Tenenbaum}, {and} \bibinfo{person}{Bill Freeman}.} \bibinfo{year}{2018}\natexlab{a}.
\newblock \showarticletitle{Visual object networks: Image generation with disentangled 3D representations}.
\newblock \bibinfo{journal}{\emph{Advances in neural information processing systems}}  \bibinfo{volume}{31} (\bibinfo{year}{2018}).
\newblock


\bibitem[Zhu et~al\mbox{.}(2018b)]%
        {r99}
\bibfield{author}{\bibinfo{person}{Jun-Yan Zhu}, \bibinfo{person}{Zhoutong Zhang}, \bibinfo{person}{Chengkai Zhang}, \bibinfo{person}{Jiajun Wu}, \bibinfo{person}{Antonio Torralba}, \bibinfo{person}{Josh Tenenbaum}, {and} \bibinfo{person}{Bill Freeman}.} \bibinfo{year}{2018}\natexlab{b}.
\newblock \showarticletitle{Visual object networks: Image generation with disentangled 3D representations}.
\newblock \bibinfo{journal}{\emph{Advances in neural information processing systems}}  \bibinfo{volume}{31} (\bibinfo{year}{2018}).
\newblock


\end{thebibliography}
\end{document}